\documentclass[letterpaper, 10 pt, conference]{ieeeconf}
\IEEEoverridecommandlockouts
\usepackage{graphicx} % for pdf, bitmapped graphics files
\usepackage{amsmath} % assumes amsmath package installed
\usepackage{amssymb}  % assumes amsmath package installed
\usepackage{multirow}
\usepackage{booktabs}
\usepackage[colorlinks,urlcolor=black,linkcolor=black,anchorcolor=black,citecolor=black]{hyperref}

\title{\LARGE \bf
ORBBuf: A Robust Buffering Method for Remote Visual SLAM}

\author{Yu-Ping Wang$^{1}$, Zi-Xin Zou$^{1}$, Cong Wang$^{1}$, Yue-Jiang Dong$^{1}$, Lei Qiao$^{2}$, Dinesh Manocha$^{3}$ \\
\\
The source code and video of applications can be found at 
\\
\url{http://github.com/Jrdevil-Wang/ORBBuf}.% <-this % stops a space
\thanks{$^{1}$Yu-Ping Wang, Zi-Xin Zou, Cong Wang and Yue-Jiang Dong are with the Department of Computer Science and Technology, Tsinghua University, Beijing, China. }%
\thanks{$^{2}$Lei Qiao is with the Beijing Institute of Control Engineering, Beijing, China. }%
\thanks{$^{3}$Dinesh Manocha is with the Department of Computer Science, University of Maryland, MD 20742, USA. }%
\thanks{Yu-Ping Wang is the corresponding author, e-mail: wyp@tsinghua.edu.cn.}%
}

\begin{document}

\maketitle
\thispagestyle{empty}
\pagestyle{empty}

\begin{abstract}

The data loss caused by unreliable network seriously impacts the results of remote visual SLAM systems.
From our experiment, a loss of less than 1 second of data can cause a visual SLAM algorithm to lose tracking.
We present a novel buffering method, ORBBuf, to reduce the impact of data loss on remote visual SLAM systems.
We model the buffering problem as an optimization problem by introducing a similarity metric between frames.
To solve the buffering problem, we present an efficient greedy-like algorithm to discard the frames that have the least impact on the quality of SLAM results.
We implement our ORBBuf method on ROS, a widely used middleware framework.
Through an extensive evaluation on real-world scenarios and tens of gigabytes of datasets, we demonstrate that our ORBBuf method can be applied to different state-estimation algorithms (DSO and VINS-Fusion), different sensor data (both monocular images and stereo images), different scenes (both indoor and outdoor), and different network environments (both WiFi networks and 4G networks).
Our experimental results indicate that the network losses indeed affect the SLAM results, and our ORBBuf method can reduce the RMSE up to 50 times comparing with the Drop-Oldest and Random buffering methods.

\end{abstract}

\section{Introduction}\label{sec:introduction}

Visual simultaneous localization and mapping (SLAM) is an important research topic in robotics~\cite{DBLP:journals/ijrr/WhelanKJFLM15}, computer vision~\cite{DBLP:conf/cvpr/ChoiZK15} and multimedia~\cite{DBLP:conf/mm/ZhangWLH19}.
The task of self-localization and mapping is computationally expensive, especially for embedded devices with both power and memory restrictions.
Remote visual SLAM systems that perform these computations on a centralized server can overcome these limitations~\cite{DBLP:conf/icra/OpdenboschOGAS18}.
On the other hand, many applications use one or more robots for specific tasks, including 3D scene reconstruction~\cite{DBLP:journals/tog/Dong0ZTXNC19} and landscape exploring~\cite{DBLP:conf/icra/JamiesonHG20}.
The centralized server in remote visual SLAM systems can be used to perform computations and collect visual information from one or more robots.
Therefore, remote visual SLAM has become an emerging research topic~\cite{DBLP:journals/vrih/ZouTY19}.

\begin{figure*}[htb]
\centering
\includegraphics[width=0.8\textwidth]{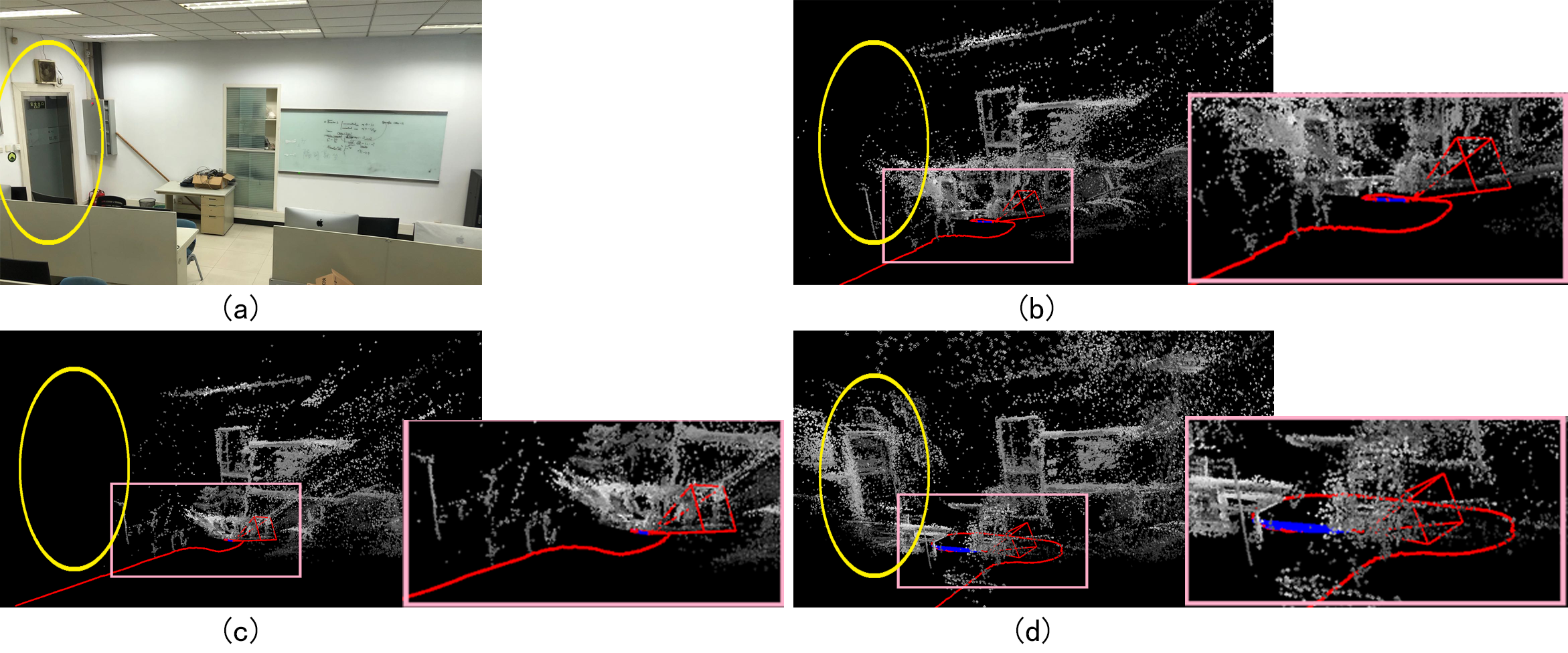}
\caption{
We highlight the benefits of our novel buffering methods (ORBBuf) for remote SLAM application.
The results of a real-world SLAM experiment using our ORBBuf algorithm with a TurtleBot3 and a server connected via a WiFi network.
(a) A view of our laboratory used for the experiment evaluation. (b) The server received visual data from the robot with ROS (which employs the Drop-Oldest buffering method~\cite{ros}) and ran a SLAM algorithm, but failed due to WiFi unreliability (the door on the left was completely missing, shown in the yellow ellipse). (c) It also failed when employing the Random buffering method~\cite{DBLP:conf/mass/EzifeLY17}. (d) By employing our ORBBuf method, the SLAM algorithm successfully estimated the correct trajectory (the red curve) and built a sparse 3D map of the scene in a reliable manner (shown as the white points).}
\label{fig:feature}
\end{figure*}

In remote visual SLAM systems, robots transmit the collected visual data (e.g. images or 3D point clouds) to a high-performance server.
This requires high network bandwidth and network reliability~\cite{DBLP:journals/jfr/GTSL16}.
For example, transmitting uncompressed 1080p images at 30 fps (frames per second) requires 1.4Gbps network bandwidth, while commodity WiFI routers can only provide 54Mbps bandwidth at most.
There has been work on reducing the bandwidth requirements based on video compression techniques~\cite{DBLP:conf/egh/HasselgrenA06,DBLP:journals/tip/ChenHC18}, compact 3D representations of sub-maps~\cite{DBLP:conf/icip/KimH07,DBLP:journals/ral/KahlerPVM16}, or down-sampling the RGB-D images~\cite{DBLP:journals/tvcg/GolodetzCLPMT18}, though few focus on addressing network reliability issues in this context.

Network connections, especially wireless ones (e.g. over WiFi or 4G), are not always reliable.
A detailed measurement study~\cite{DBLP:journals/icl/HooftPWHRBT16} has shown that lower throughput or even network interruption could occur for dozens of seconds due to tunnels, large buildings, or poor coverage in general.
With the advent of 5G, the problems of network bandwidth and latency will be relieved, but the unreliability due to poor coverage still exists~\cite{DBLP:journals/twc/ChenGN19}.

In this paper, we address that the problem of network reliability can have considerable impact on the accuracy of remote visual SLAM systems, and our approach is orthogonal to the methods that reduce the bandwidth requirements.
To verify our claim, we built a remote visual SLAM system that connects a Turtlebot3 and a server via public WiFi router.
The robot moved around our laboratory as shown in Figure~\ref{fig:feature}(a).
We fixed a camera on top of the robot, and the captured images were transmitted to a server via a public WiFi router.
We found that the SLAM algorithm~\cite{DBLP:journals/pami/EngelKC18} on the server repeatedly failed (highly inaccurate results or even lost tracking) at some certain locations.
At these locations, the network connection was highly unreliable (it may have been affected by the surrounding metal tables).
To tolerate such network unreliability, a common solution is buffering.
When the network is unreliable, the robot puts new frames into its buffer and waits for future transmission.
With such unreliability persists, the buffer becomes full and the buffering method is responsible for deciding which frame(s) should be discarded.
We tried two kinds of commonly used buffering methods (Drop-Oldest and Random~\cite{DBLP:conf/mass/EzifeLY17}), but the SLAM algorithm failed in both cases (shown in Figure~\ref{fig:feature}(b) and (c)).

\textbf{Main Results:} In this paper, we present a novel robust buffering method, \textit{ORBBuf}, for remote visual SLAM.
By analyzing the requirement of visual SLAM algorithms, we introduce a \textbf{\textit{similarity}} metric between between frames and model the buffering problem into an optimization problem that tries to maximize the minimal similarity.
We use an efficient greedy-like algorithm, and our ORBBuf method drops the frame that results in the least impact to visual SLAM algorithms.
Experimental results on real-world scenarios (shown in Figure~\ref{fig:feature}(d)) and tens of gigabytes of 3D visual datasets show that the network interruptions indeed affect the SLAM results and that our ORBBuf method can greatly relieve the impacts. The main contributions of this paper include:

\begin{itemize}
    \item We address the problem of network reliability for remote visual SLAM. We model the buffering problem into an optimization problem by introducing the notion of similarity. We propose a novel buffering method named ORBBuf to solve this problem.
    \item We conduct studies that reveal how much SLAM algorithms rely on the correlations between the consecutive input frames and use that studies to propose a novel similarity metric.
    \item We implement our ORBBuf method based on a widely used middleware framework (ROS~\cite{ros}), and it will be open-sourced for the benefit of the community.
    \item Through an extensive evaluation on real-world scenarios and tens of gigabytes of 3D vision datasets, we demonstrate that our ORBBuf method can be applied to different state-estimation algorithms (VinsFusion~\cite{DBLP:journals/trob/QinLS18} and DSO~\cite{DBLP:journals/pami/EngelKC18}), large outdoor (KITTI~\cite{DBLP:journals/ijrr/GeigerLSU13}) and indoor (TUM monoVO~\cite{DBLP:journals/corr/EngelUC16}) datasets, and different network environments (WiFi network and 4G network~\cite{DBLP:journals/icl/HooftPWHRBT16}). The experimental results show that using our ORBBuf method can reduce the RMSE (root mean square error) up to 50 times comparing with the Drop-Oldest and Random buffering methods.
\end{itemize}

\section{Related Work}\label{sec:related}

\subsection{Visual SLAM}

Visual information is usually large in size.
To lower the computational requirement, feature-based approaches extract feature points from the original images and only use these feature points to decide the relative pose between frames through matching.
A typical state-of-the-art feature-based visual SLAM is ORB-SLAM~\cite{DBLP:journals/trob/Mur-ArtalMT15,DBLP:journals/trob/Mur-ArtalT17}.
The most recent version of ORB-SLAM3~\cite{DBLP:journals/corr/abs-2007-11898} has combined most of the visual SLAM techniques into a single framework based on the ORB~\cite{DBLP:conf/iccv/RubleeRKB11} feature.
Accurate localization depends on the number of successful matched feature points among consecutive frames.
The correlations among consecutive frames are not robust~\cite{DBLP:journals/trob/CadenaCCLSN0L16}.
When too many consecutive frames are lost, visual SLAM algorithms will fail to match enough corresponding feature points, leading to inaccurate results or even failure of the system.
We will verify this claim with a more detailed analysis in Section~\ref{subsec:choice}.
In this case, the loop-closure routine, which usually works as a background optimization to relieve accumulation errors, may save the SLAM system from a total failure, but it is both expensive and unstable~\cite{DBLP:journals/jfr/LabbeM19}.
In addition to feature-based approaches, there are direct SLAM algorithms that use all raw pixels, including DTAM~\cite{DBLP:conf/iccv/NewcombeLD11}, LSD-SLAM~\cite{DBLP:conf/eccv/EngelSC14}, and DSO~\cite{DBLP:journals/pami/EngelKC18}.
They try to minimize the photometric error between frames by probabilistic models directly using the raw pixels.
The success of all these SLAM algorithms depends on the correlations between consecutive frames, which indicate the requirements for SLAM algorithms.
In this paper, we introduce fast feature extraction into a buffering method to relieve the impact of network unreliability on SLAM algorithms.

\subsection{Remote Visual SLAM}

Remote visual SLAM systems adopt a centralized architecture, as in PTAMM~\cite{DBLP:conf/iswc/CastleKM08}, C2TAM~\cite{DBLP:journals/ras/RiazueloCM14}, and CVI-SLAM~\cite{DBLP:journals/ral/KarrerSC18}.
These methods are easily implementable compared with the emerging decentralized architecture~\cite{DBLP:conf/icra/CieslewskiCS18}.
A major issue for remote visual SLAM is what data representation should be transmitted.
Compress-then-analyze (CTA) solutions transmit compressed sensor data such as RGB images and depth images~\cite{DBLP:journals/tvcg/GolodetzCLPMT18}.
Analyze-then-compress (ATC) solutions assume that robots are powerful enough to build a local map or generate hints of the local map, and then share them among robots or merge them at the server~\cite{DBLP:journals/tog/Dong0ZTXNC19}.
Opdenbosch et al.~\cite{DBLP:conf/icra/OpdenboschOGAS18} proposed an intermediate solution.
Fast feature extraction is performed at the robot.
These features are collected by the server and the server performs the following SLAM task.
These solutions have a big impact on bandwidth requirement, but are orthogonal with the network reliability problem.
In Section~\ref{sec:evaluation}, our evaluations employ the CTA solutions since
our robots are not powerful enough and we will show that adjusting compression parameters cannot solve the network reliability problem of remote visual SLAM.

On the other hand, few works have focused on network reliability.
Many works explicitly assume that the network bandwidth is sufficient and stable~\cite{DBLP:conf/icra/KimKFLBRT10,DBLP:conf/icra/CieslewskiCS18,DBLP:journals/ral/LajoieRCCB20}.
This assumption is hard to satisfy in real-world wireless networks~\cite{DBLP:journals/icl/HooftPWHRBT16,DBLP:journals/twc/ChenGN19}.
Decentralized multi-robot SLAM systems~\cite{DBLP:conf/icra/CarloneNDBI10} usually consider scenarios where the robots only occasionally encounter each other, but they assume that the memory used for buffering is large enough.
In this paper, we address the problem of network reliability and propose a novel buffering method that is robust against network unreliability with a limited buffer size.

\subsection{Buffering Methods in Other Applications}

Buffering methods are widely used in other applications such as video streaming and video conferencing.
Dynamic Adaptive Streaming over HTTP (DASH)~\cite{DBLP:conf/mmsys/Stockhammer11} has become the de facto standard for video streaming services.
By using the DASH technique, each video is partitioned into segments and each segment is encoded in multiple bitrates.
Adaptive BitRate (ABR) algorithms decide which bitrate should be used for the next segment.
However, the DASH technique is not quite suitable for remote visual SLAM for two main reasons.
First, visual SLAM algorithms rely on correlations between consecutive frames.
Using multiple bitrates potentially weakens such correlations and increases errors.
Second, correlations among consecutive frames are not robust~\cite{DBLP:journals/trob/CadenaCCLSN0L16}.
Since the DASH technique partitions video into segments, losing even a single segment would lead to failure.

In addition to the DASH technique, buffering methods are common solutions against network unreliability and have been researched for the last decade.
Specialized for video transmission, enhanced buffering methods introduce the concept of \textit{profit} and assume that each message has a constant profit value.
The goal is to maximize the total profit of the remaining frames~\cite{DBLP:journals/sigact/Goldwasser10}.
According to the application scenarios, different types of frames may have different profits~\cite{DBLP:conf/secon/KrifaBS08,DBLP:conf/csa2/KimW17,DBLP:journals/cn/Scalosub17,DBLP:journals/pomacs/YangWH17,DBLP:conf/apcc/KimuraM17}.
However, in the visual SLAM scenario, the profit is not a constant value for each frame but is rather a correlation among its neighbors.
Therefore, we need better techniques that understand the requirements of visual SLAM algorithms to correlate the input frame sequences.

\section{Our Approach: ORBBuf}\label{sec:approach}

%In this section, we model the buffering problem as an optimization problem by introducing the notion of similarity value between frames (Section~\ref{subsec:buffer}).
%We then solve this problem with our ORBBuf method (Section~\ref{subsec:orbbuf}).
%and describe its implementation (Section~\ref{subsec:implementation}).
%Finally, we discuss why alternative buffering methods fail and our introduced notion of similarity value can overcome the problem (Section~\ref{subsec:choice}).

\subsection{Notations}\label{subsec:notation}

For clarity, we define some notations and symbols that we will use to model the buffering problem.

\begin{itemize}
    \item $S$ denotes a message sequence. Messages in the message sequence $S$ are denoted as $S_1, S_2, ..., S_n$. $|S|$ denotes the number of messages of $S$, i.e. $|\{S_1, S_2, ..., S_n\}| = n$.
    \item $S^{send}$ denotes the message sequence generated at the edge-side device. Since messages are generated over time, $S^{send}$ grows correspondingly. $S^{send}(T)$ denotes the status of $S^{send}$ at time $T$.
    \item Without loss of generality, we define time $T$ as the message $S^{send}_T$ is generated, i.e., $S^{send}(T) = \{S^{send}_1, S^{send}_2, ..., S^{send}_T\}$.
    \item $S^{send*}$ denotes the message sequence that is actually sent via the network. $S^{recv}$ denotes the message sequence that is received at the server.
    \item $L$ denotes the size of the sending buffer. $B(T)$ denotes the buffer status at time $T$.
\end{itemize}

\subsection{Design of Similarity Metric}

As we have stated in Section~\ref{sec:related}, the success of SLAM algorithms depends on the correlations between consecutive frames, i.e. $S^{recv}$.
In this paper, we define the correlations between two frames as their \textbf{\textit{similarity}} and denote the similarity function as $M(S_1, S_2)$.
Larger similarity value leads to more robust SLAM results.
We formalize this prior knowledge as Equation~(\ref{equ:similarity}).
The probability of the success at time $t$ of the SLAM algorithm is 	proportional to the similarity between two adjacent frames at time $t$.

\begin{equation}
\label{equ:similarity}
    P_{Success}(t) \propto M(S^{recv}_{t}, S^{recv}_{t + 1})
\end{equation}

In fact, different SLAM algorithms use different models to represent similarity, but they all reflect the correlation between the frames themselves.
For example, the ORB-SLAM algorithm~\cite{DBLP:journals/trob/Mur-ArtalMT15}  is based on the ORB feature.
ORB feature refers to ``Oriented FAST and Rotated BRIEF" and corresponds to a very fast binary descriptor based on BRIEF, which is rotation invariant and resistant to noise~\cite{DBLP:conf/iccv/RubleeRKB11}.
During its tracking routine, ORB feature points are extracted from each frame and matched between consecutive frames.
Insufficient matched feature points leads to the failure of tracking and even the failure of the SLAM algorithm.
On the contrary, more matched feature points leads to smaller error of the tracking results and the SLAM results.
Without loss of generality, we use the number of successfully matched ORB feature points to measure the similarity because ORB features can be calculated efficiently.

\subsection{Modeling the Buffering Problem}\label{subsec:buffer}

Messages containing 3D visual information are usually large in size.
However, network packets are limited in size (e.g., 1500 bytes), meaning that these messages are divided into hundreds of packets.
If each message contains 100 packets (e.g., a typical JPEG compressed image), losing even 1\% of the packets means about $1-(1-0.01)^{100} \approx 63.4\%$ of the messages would not be complete.
Therefore, unreliable network protocols such as UDP are not preferred in this case.
When using reliable network protocols (e.g., TCP) or specialized protocols (e.g., QUIC~\cite{quic}), a network connection is dedicated to sending a single message at a time while other messages are inserted into a sending buffer to wait for transmission.
Therefore, we can assume that all messages that are actually sent via the network are received by the server, i.e. $S^{send*} = S^{recv}$.
When the network bandwidth is insufficient or a temporary network interruption occurs, the sending buffer becomes full.
In this case, the buffering method $P$ has to drop some messages from the sending buffer, which decides $S^{send*} \subseteq S^{send}$.
As we can see, $S^{recv} = S^{send*} \subseteq S^{send}$ is decided by the network status $D$ and the buffering method $P$.
Finally, all messages in $S^{recv}$ are provided to the SLAM algorithm.

Our goal is to keep the SLAM algorithm robust.
Since larger similarity value leads to more robust SLAM results, we can state the buffering problem as follows.

\textbf{Problem Statement}: Knowing the buffer size $L$ while not knowing the transmission status $D$ or the message sequence $S^{send}$ in advance, produce a buffering method $P$ that maximizes the multiplication of the similarities over adjacent messages in $S^{recv}$. Described as Equation (\ref{equ:problem}).

\begin{equation}
\label{equ:problem}
    P = \arg\max (\prod_{t} M(S^{recv}_{t}, S^{recv}_{t+1}))
\end{equation}

This equation is a simple accumulation of Equation~(\ref{equ:similarity}).
From the problem statement, we can see that the key to solving this problem is understanding the similarity between frames.

\subsection{Our ORBBuf Method}\label{subsec:orbbuf}

As we have stated in Equation (\ref{equ:problem}), our goal is to maximize the minimal similarity between adjacent frames in $S^{recv}$.
The only information provided to a buffering method $P$ is the buffer status $B(T)$ and the next message $S^{send}_T$.
Therefore, we cannot get the global optimal solution.
However, we could use a greedy-like algorithm to solve each local problem.
Since $S^{recv}$ is combined with the output from buffer $B$, we can define a simplified local problem as Equation (\ref{equ:local}).
Our ORBBuf method finds a result $B'$ that maximizes the minimal similarity of all feasible $B'$.

\begin{equation}
\label{equ:local}
    P_{ORBBuf}(B, S_T) = \arg\max (\min_{t} M(B'_{t}, B'_{t+1}))
\end{equation}

To get the optimal solution of this local problem efficiently, we record a score for each message, which is the similarity between its \textit{previous} message and the \textit{next} message in the buffer.
The score of a message indicates the resulting adjacent similarity when it is dropped.
When the buffer is full, our ORBBuf method will drop the message with the maximal score.
Since the similarity between other adjacent messages remains the same, the resulting minimal similarity relies on whether the score is lower than the original minimal similarity.

\begin{figure}[htb]
\centering
\includegraphics[width=0.8\columnwidth]{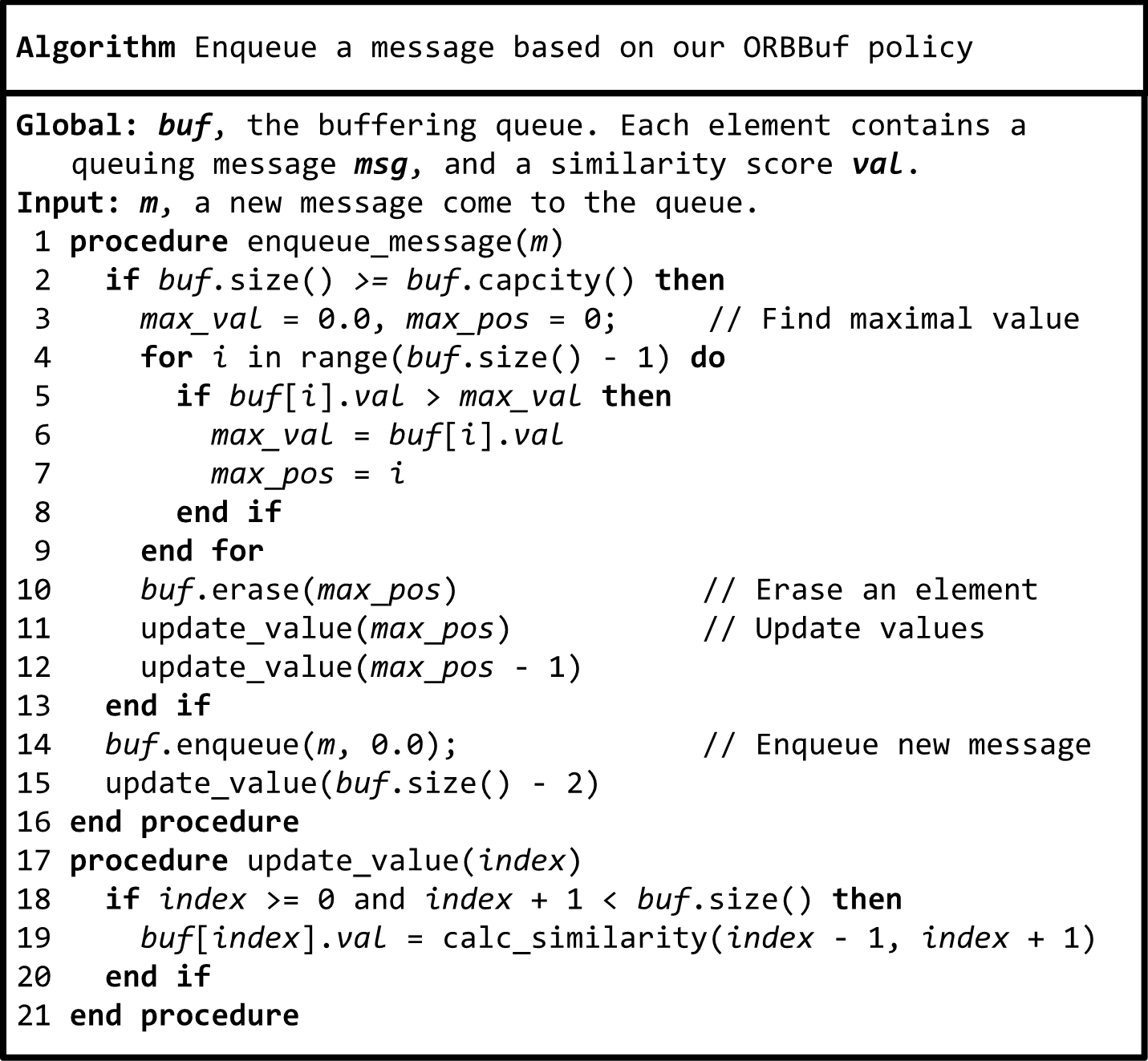}
\caption{The pseudo code of our ORBBuf method. It is a greedy-like algorithm that maximizes the resulting minimal adjacent similarity within O(1) time. }
\label{fig:algorithm}
\end{figure}

The pseudo code of our ORBBuf method is shown in Figure~\ref{fig:algorithm}.
From line 3 to line 9, we find the message with the maximal score.
This message is dropped in line 10.
Dropping this message causes the score of some messages to be changed in lines 11 and 12.
After the new message is inserted in line 14, the score of the previous message is changed in line 15.
Note that the function $update\_value$ is called at most 3 times each time a message arrives at the buffer.
Since the calculation of the ORB feature is fast, the overhead introduced by our ORBBuf method is negligible.
We will verify this claim in Section~\ref{sec:evaluation}.

Overall, with the cost of the memory for recording a score for each message (which is tiny compared with the size of each message), our ORBBuf method can find the solution within O(1) time, which is independent of the buffer size.

\subsection{Implementation}\label{subsec:implementation}

We implement our ORBBuf method based on ROS~\cite{ros}, which is a middleware widely used in robotic systems.
The interface of ROS does not notify the upper-level application when the lower-level transmission is completed.
Thus, we add a notification mechanism and an interface to ROS.
At runtime, the new message is put into a buffer following our ORBBuf method, and the low-level transmission is evoked, if possible.
Whenever the low-level transmission is completed, we begin transmission of another message.
We modify the code of the class \textit{TransportSubscriberLink}, located at the \textit{ros\_comm} project~\cite{ros_comm}, and link it into \textit{libroscpp.so}.
Developers can easily call on our interface to send a new message with our ORBBuf method.

\subsection{Analysis and Comparison}\label{subsec:choice}

Before discussing our quantitative evaluation, we would like to compare two alternative buffering methods by analyzing the characteristics of visual SLAM algorithms.

A commonly-used but trivial buffering method is the \textit{Drop-Oldest} method.
It is designed based on the assumption that newer messages are more valuable than older messages.
Therefore, if the buffer is full when a new message arrives, the oldest message is dropped to make room for the new message (formalized in Equation (\ref{equ:dropoldest})).
As a result, $S^{recv}$ will directly reflect the network status $D$.
When message transmission is slower than message generation, the buffer eventually becomes full, and consecutive frames would be dropped during network interruption.

\begin{equation}
\label{equ:dropoldest}
\begin{split}
    P_{Drop-Oldest}(\{B_1, B_2, ..., B_l\}, S_T) = & \\
    \begin{cases}
        \{B_1, B_2, ..., B_l, S_T\}, & l < L, \\
        \{B_2, B_3, ..., B_L, S_T\}, & l = L.
    \end{cases}
\end{split}
\end{equation}

\begin{figure}[htb]
\centering
\includegraphics[width=\columnwidth]{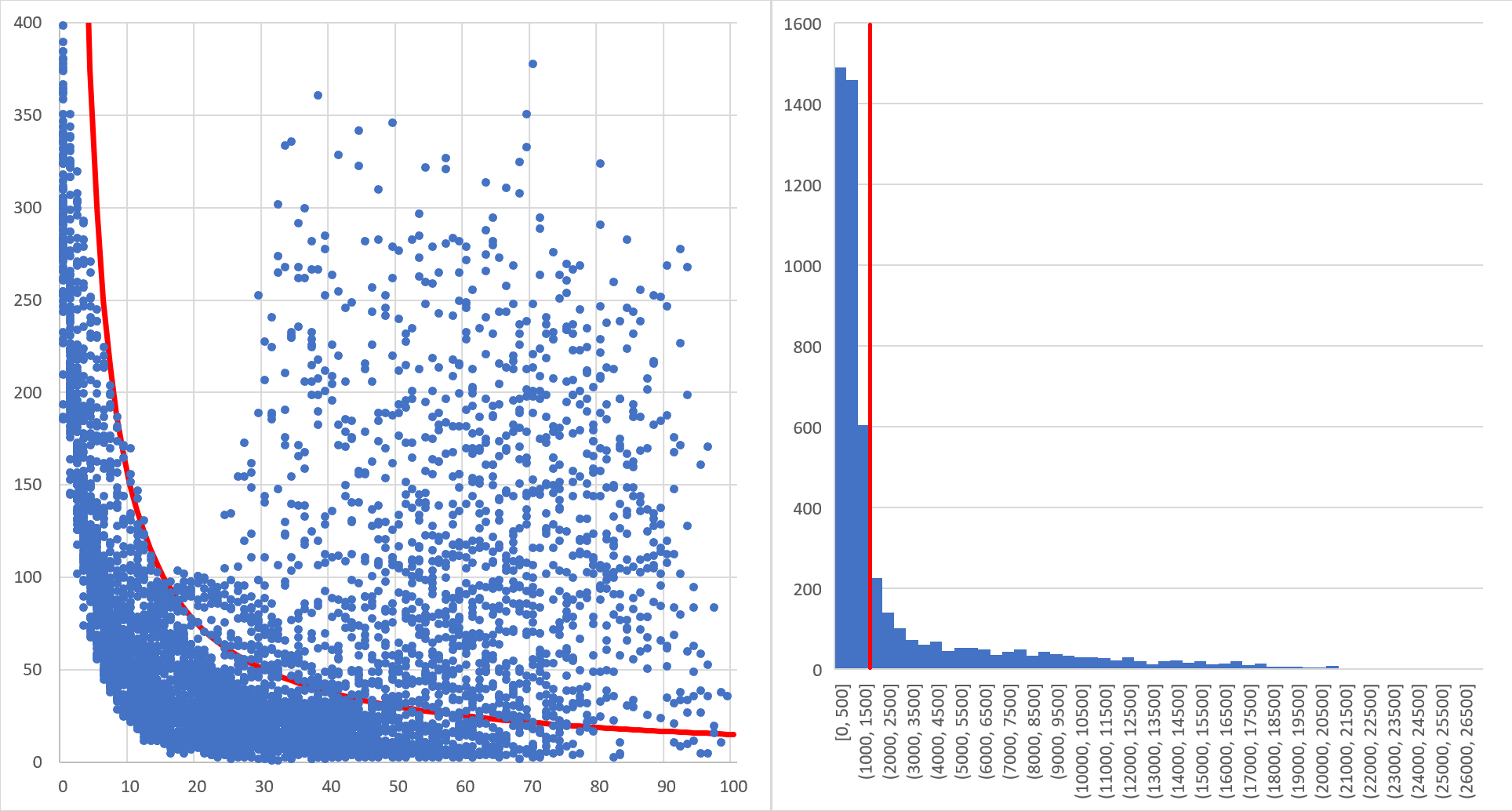}
\caption{Relationship between the frame distance and the similarity. 70.3\% of points are under the red curve on the left and they show a coarse inverse relationship between the frame distance and the similarity.}
\label{fig:correlation}
\end{figure}

To show the consequence of losing consecutive frames, we conduct a study with sequence No. 11 of the TUM monoVO dataset~\cite{DBLP:journals/corr/EngelUC16} and pick the frames from No. 200 to No. 300.
For every two frames, we record their distance (the difference between their frame numbers) and their similarity.
The results are plotted in Figure~\ref{fig:correlation}.
At the left of Figure~\ref{fig:correlation}, the horizontal axis is the distance between two frames and the vertical axis is their similarity.
We can clearly see that there is a coarse inverse relationship between the frame distance and the similarity.
We note that there are more exceptions when the distance becomes larger, which is reasonable since there may be similar objects in an indoor scene.
In order to show the inverse relationship more clearly, we draw the corresponding histogram at the right of Figure~\ref{fig:correlation}.
The horizontal axis is the product of the distance and similarity and the vertical axis is the number of pairs of frames within the interval.
We note that 70.3\% of the pairs are within the first 3 intervals (from 0 to 1500), which is shown as the red curve (on the left) or the red line (on the right).
Overall, we can conclude that there is a coarse inverse relationship between the frame distance and the similarity.
Therefore, when consecutive frames are dropped by the Drop-Oldest method, a large frame distance occurs, and we are thus more likely to get a lower similarity and unstable SLAM results.

Similarly, another simple buffering methods, Drop-Youngest (the youngest message is dropped, formalized in Equation (\ref{equ:dropyoungest})), also drops consecutive frames during network interruption.
The only difference is whether the oldest $L$ messages are retained (Drop-Youngest) or the youngest $L$ messages are retained (Drop-Oldest).
Therefore, when using the Drop-Youngest method, we are also likely to get a lower similarity and unstable SLAM results.

\begin{equation}
\label{equ:dropyoungest}
\begin{split}
    P_{Drop-Youngest}(\{B_1, B_2, ..., B_l\}, S_T) = & \\
    \begin{cases}
        \{B_1, B_2, ..., B_l, S_T\}, & l < L, \\
        \{B_1, B_2, ..., B_{L-1}, S_T\}, & l = L.
    \end{cases}
\end{split}
\end{equation}

A promising buffering method to avoid losing consecutive frames is the \textit{Random} method.
If the buffer is full when a new message arrives, this method randomly drops a message inside the buffer (formalized in Equation (\ref{equ:randomdrop}), where $r$ is a random number between $1$ and $L$).
Statistically, the random method drops the messages in a uniform manner.
Such uniform frame distance is more likely to result in average similarity and solve the problem formalized in Equation (\ref{equ:local}).

\begin{equation}
\label{equ:randomdrop}
\begin{split}
    P_{Random}(\{B_1, B_2, ..., B_l\}, S_T) = & \\
    \begin{cases}
        \{B_1, B_2, ..., B_l, S_T\}, & l < L, \\
        \{B_1, B_2, ..., B_{r-1}, B_{r+1}, ..., B_L, S_T\}, & l = L.
    \end{cases} \\
\end{split}
\end{equation}

In practice, this method cannot solve our buffering problem for two reasons.
First, as we have stated in Equation (\ref{equ:problem}), we need to maximize the minimal similarity, which is not even statistically guaranteed.
Second, similarity is not distributed in a uniform manner.

\begin{figure}[htb]
\centering
\includegraphics[width=\columnwidth]{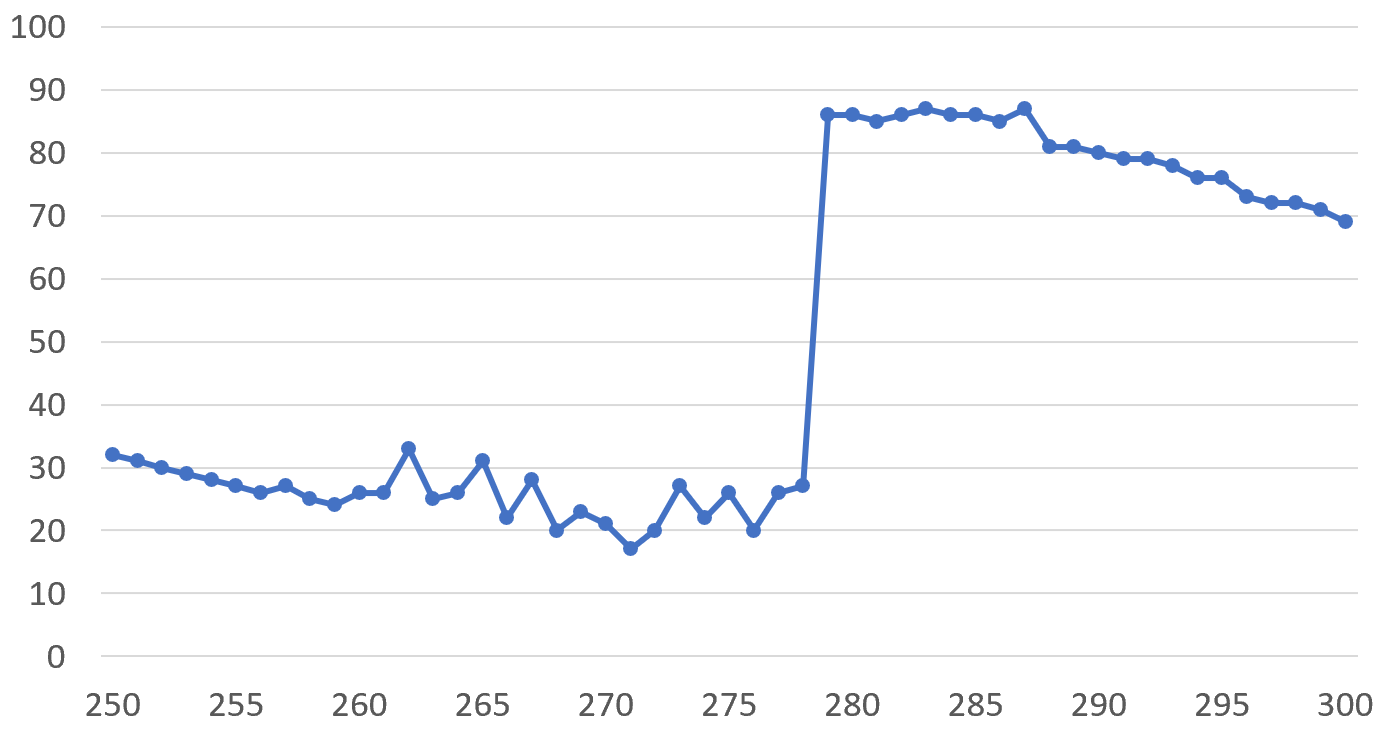}
\caption{Simulation results when a SLAM algorithm loses consecutive frames. The SLAM algorithm can tolerate some lost frames, but the tolerance capability is uneven at different positions (less than 1 second in the worst case at frame 271).}
\label{fig:stable}
\end{figure}

To verify this claim, we conduct another study with the same data sequence that we used in the first study.
We remove some interval of frames from the original sequence, provide it to a SLAM algorithm (the DSO~\cite{DBLP:journals/pami/EngelKC18} algorithm), and record whether the algorithm would fail.
The results are shown in Figure~\ref{fig:stable}.
The horizontal axis is the number of the first lost frame, and the vertical axis is the number of consecutive frames that would cause the SLAM algorithm to fail.
For example, the value at 250 is 32, which means the algorithm failed when the frames from 250 to 282 were lost, but the algorithm still succeeded when the frames from 250 to 281 were lost.
In the worst case (at frame 271), losing 17 consecutive frames (at 25 fps, i.e. 0.68 seconds) would cause the algorithm fail.
From this result, we can see that the SLAM algorithm can tolerate some lost frames, but the tolerance capability is uneven at different positions.

~

Overall, these two solutions cannot solve our problem.
They fail because they are general-purpose methods and do not take the equirements of SLAM algorithms into consideration.
In contrast, our ORBBuf method employs similarity to guide dropping procedures.
In our evaluation, we will compare our ORBBuf method with these two methods.

\section{Evaluation}\label{sec:evaluation}

To show the advantage of our ORBBuf method, we carry out three experiments.
In these experiments, different kinds of network situations, different SLAM algorithms, and different kinds of input data are employed.

To show practical results, our experiments are carried out completely online on real-world hardware.
We use a laptop with an Intel Core i7-8750H @2.20GHz 12x CPU, 16GB memory, and GeForce GTX 1080 GPU as the SLAM server.
In the first two experiments, large vision datasets are used.
We use a laptop without GPU support to replay each data sequence and transmit it to the server.
All software modules are connected with the ROS middleware.
The ROS version is Lunar on Ubuntu 16.04.

\begin{table*}[htb]
\centering
\caption{Numeric comparison results of different buffering methods tolerating simulated network interruption.}
\includegraphics[width=0.65\textwidth]{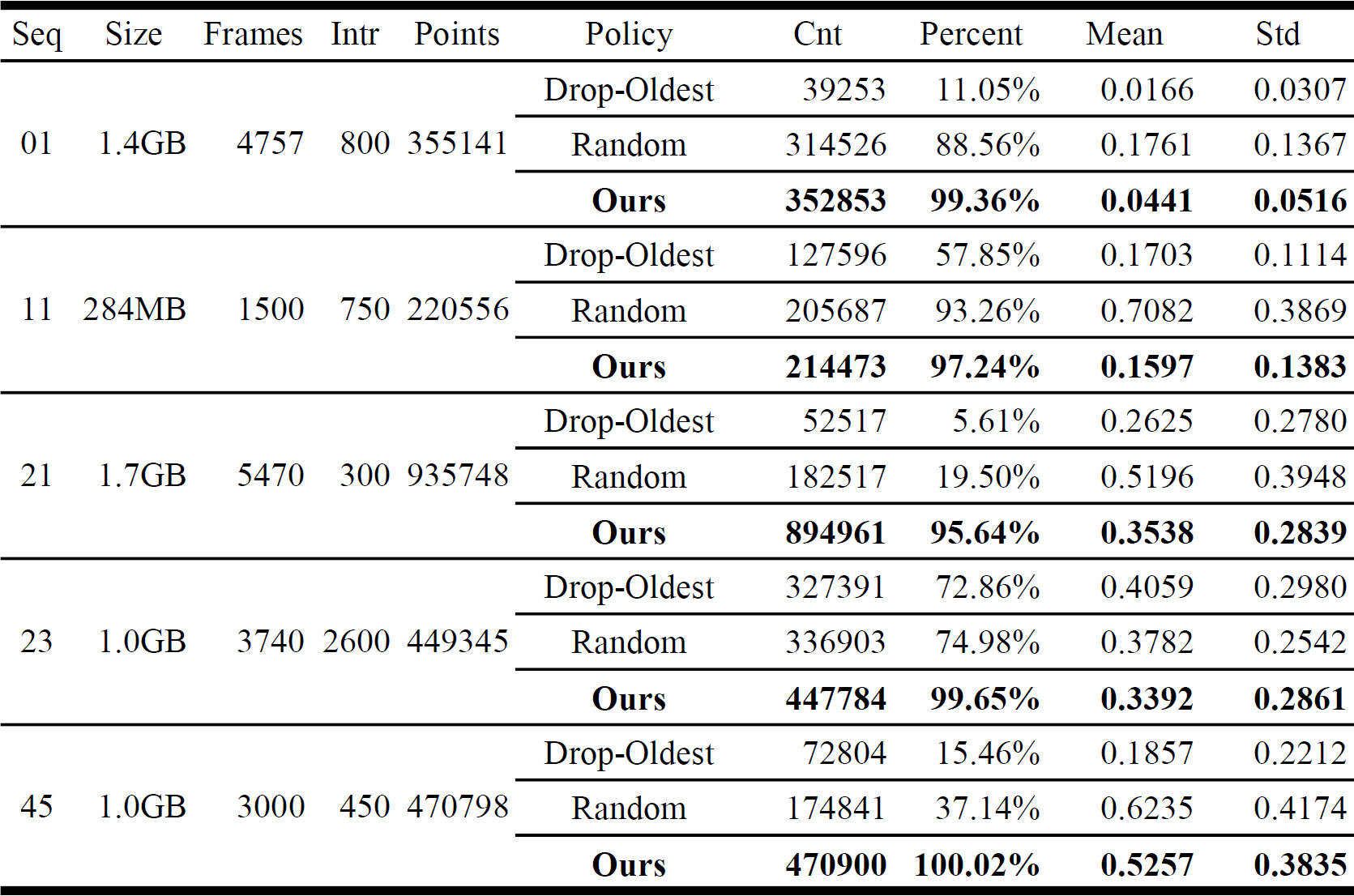}
\label{tab:reconstruction}
\end{table*}

\begin{figure*}[htb]
\centering
\includegraphics[width=\textwidth]{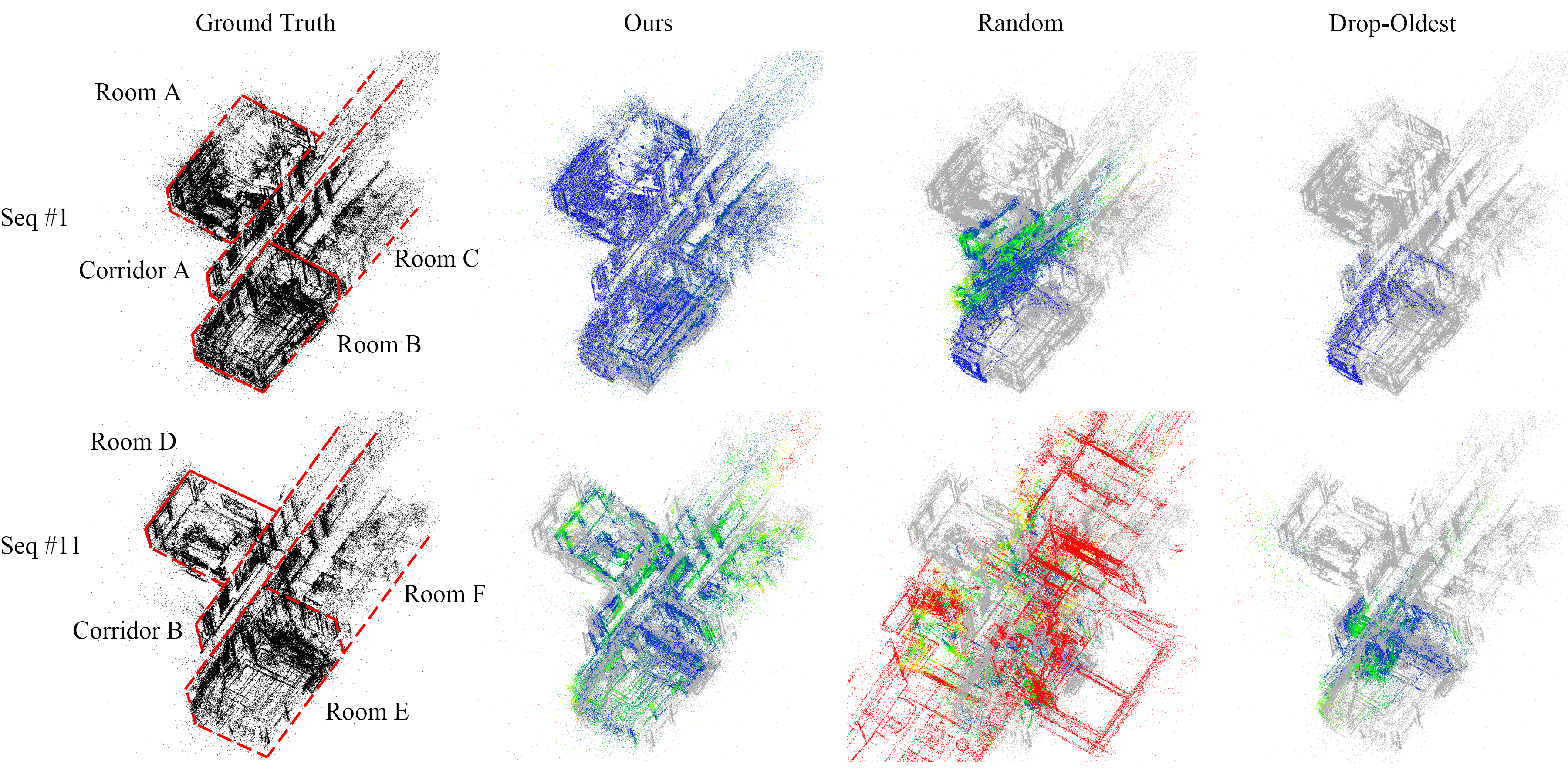}
\caption{Two visualized results of different buffering methods tolerating simulated network interruption. Sequence 1 and 11 of the TUM monoVO dataset are both collected from indoor scenes that include rooms and corridors. The comparisons are generated with CloudCompare, where grey points represent the ground truth, red points represent large errors and blue points represent small errors.}
\label{fig:reconstruction}
\end{figure*}

\subsection{Simulated Network Interruption}

In this experiment, we use the TUM monoVO~\cite{DBLP:journals/corr/EngelUC16} dataset as the input data sequence and run the DSO~\cite{DBLP:journals/pami/EngelKC18} algorithm on the SLAM server.
All data sequences are replayed at 25 fps, and the buffer size is set to one second.
The TUM monoVO dataset is a large indoor dataset that contains 50 real-world sequences (43GB in total) captured with a calibrated monocular camera. All images in this dataset are compressed as \textit{.jpg} files and have a resolution of 1280x1024.
The robot and the server are connected via a real-world 54Mbps WiFi router. To simulate network interruption, we modify the data sequence replaying module by adding three new parameters. Before transmitting a designated frame, ``\textit{Intr}", we add a \textit{T} milliseconds latency to the network using the \textit{tc} command to simulate network interruption, where the time of each network interruption lasts for \textit{L} frames. During this experiment, the network latency \textit{T} is set to 1000 milliseconds, and the network interruption duration \textit{L} is set to 50 frames.

Table~\ref{tab:reconstruction} gives numeric evaluation results. In this table, ``Seq" denotes the sequence number in the dataset, ``Size" denotes the total size of the data sequence, ``Frames" denotes the total number of frames, ``Intr" denotes the frame when network interruption occurs, ``Points" denotes the total number of points in the ground truth result, ``Policy" denotes the buffering method used, ``Cnt" denotes the number of points in the result when using the buffering method, ``Percent" denotes the completeness (Cnt divided by Points), ``Mean" denotes the mean error distance between corresponding points, and ``Std" denotes the standard deviation of the error distances.
Figure~\ref{fig:reconstruction} shows two visualized results.

As we can see, when using the Drop-Oldest method, the SLAM algorithm is most likely to fail when the network interruption occurs (the percentage is low). When using the Random method, the SLAM algorithm still fails in some cases. When using our ORBBuf method, the SLAM algorithm succeeds in all cases and the error distances between corresponding points are smaller than other successful cases.

\textbf{Effect of compression.} All results in Table~\ref{tab:reconstruction} are obtained with the default ROS image-compression configuration (JPEG with quality level 80).
To show the effect of compression, we change the JPEG quality level to 10 (lower quality for reducing image size) and re-run the experiment with sequence 11.
With this configuration, the compression algorithm can reduced the bandwidth requirement from about 3MB/s to 0.5MB/s.
When using the Drop-Oldest method, the DSO algorithm also loses the track at the network interruption.
The result point cloud includes 55382 points (25.11\%) and the mean error distance is 0.2193.
The number of points gets lower and the mean distance is higher because the lower image quality affect the extraction of feature points.
Less feature points leads to less points in the result point cloud and introduces more error due to inaccurate matching.
In conclusion, although the compression algorithm can reduce the bandwidth requirement, it cannot help tolerate network interruption and introduces more error in the mean time.

\subsection{Collected 4G Network Traces}

In this experiment, we use the KITTI~\cite{DBLP:journals/ijrr/GeigerLSU13} dataset as the input data sequence and run the VINS-Fusion~\cite{DBLP:conf/iros/QinS18,DBLP:journals/trob/QinLS18} algorithm on the server.
The KITTI dataset is a large outdoor dataset contains 22 real-world sequences (22.5GB in total) captured with calibrated stereo cameras fixed on top of an automobile.
All images in this dataset are \textit{.png} files and have a resolution of 1241x376.
During this experiment, all data sequences are replayed at 10 fps and the buffer size is set to two seconds.
We employ the network traces collected from a real-world 4G network by~\cite{DBLP:journals/icl/HooftPWHRBT16}.
The robot and the server are connected via a network cable to minimize other factors.
In~\cite{DBLP:journals/icl/HooftPWHRBT16}, dozens of network traces are collected for different scenarios.
Since the KITTI dataset is collected on an automobile, we choose four of the network traces that are also collected on a driving car.

\begin{figure*}[htb]
\centering
\includegraphics[width=\textwidth]{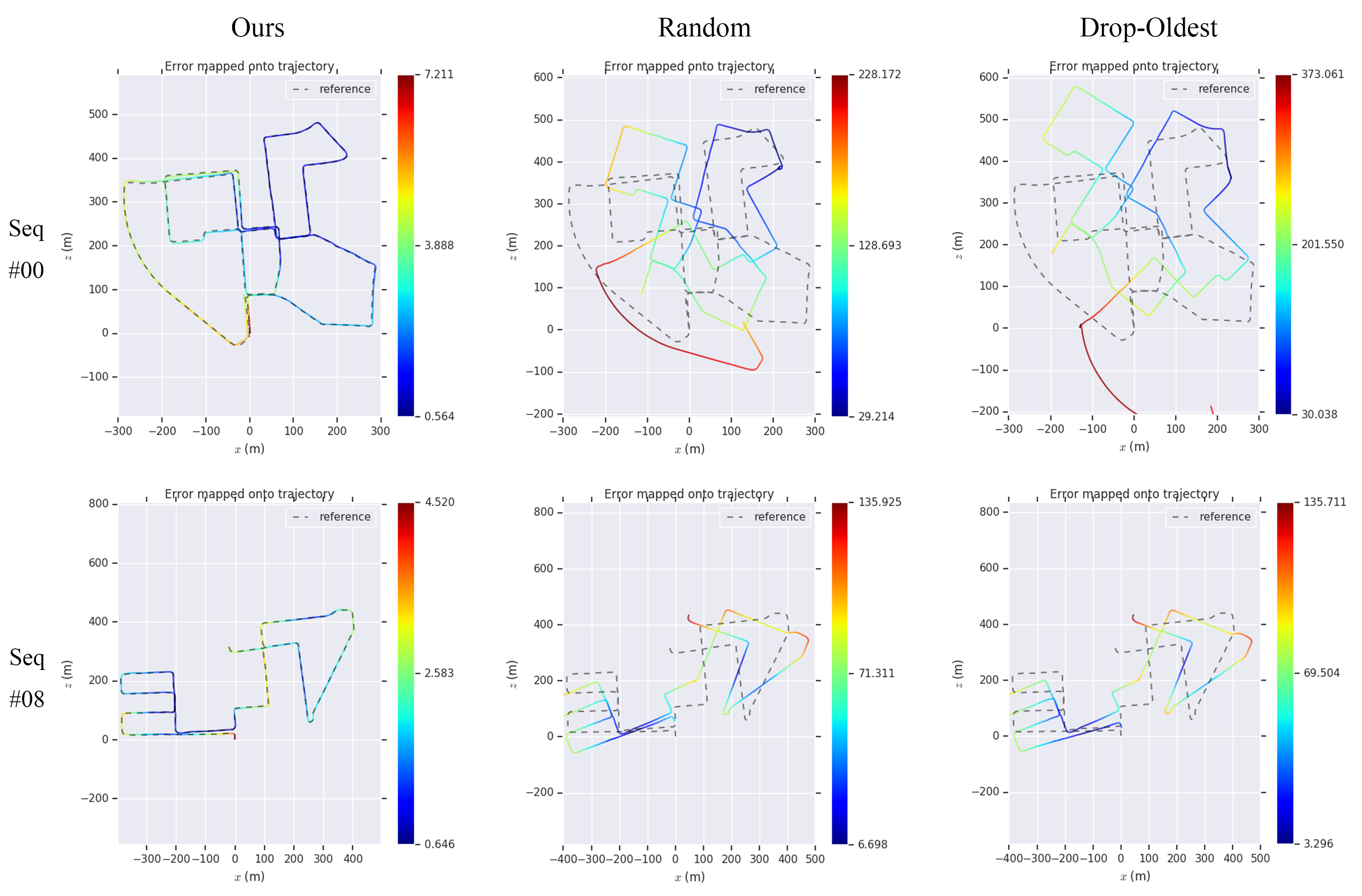}
\caption{Two visualized results of different buffering methods handling collected 4G network traces. Sequence 00 and 08 of the KITTI dataset are collected from outdoor scenes on urban roads. The comparisons are generated with the evo project~\cite{grupp2017evo}.}
\label{fig:localization}
\end{figure*}

Table~\ref{tab:localization} gives numeric evaluation results.
In this table, ``Seq" denotes the sequence number in the dataset, ``Size" denotes the total size of the data sequence, ``Frames" denotes the total number of frames, ``Net Trace" denotes the network traces, and ``RMSE" denotes the root mean square error between the ground truth and the result using a buffering method.
Figure~\ref{fig:localization} shows two visualized results.

\begin{table*}[htb]
\centering
\caption{Numeric comparison results of different buffering methods handling collected 4G network traces.}
\includegraphics[width=0.65\textwidth]{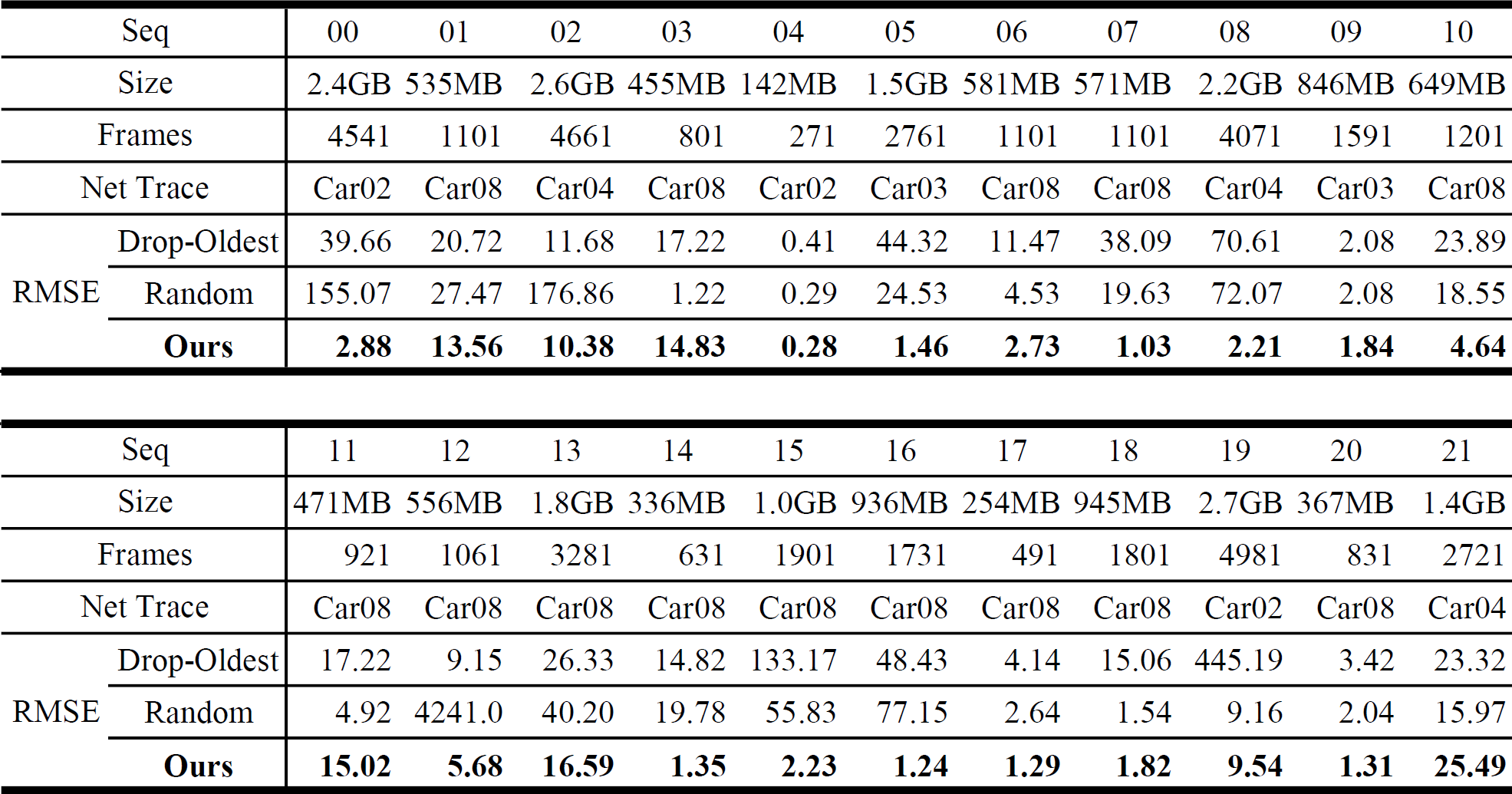}
\label{tab:localization}
\end{table*}

The VINS-Fusion algorithm never alerts a failure, but the result can be highly unstable.
As we can see, when using the Drop-Oldest method, the resulting RMSE values are relatively large.
%When using the Random method, the algorithm can not always achieve better results.
When using our ORBBuf method, the resulting trajectory fits the ground truth much better and the RMSE values are reduced by up to 50 times.

\begin{figure}[htb]
\centering
\includegraphics[width=\columnwidth]{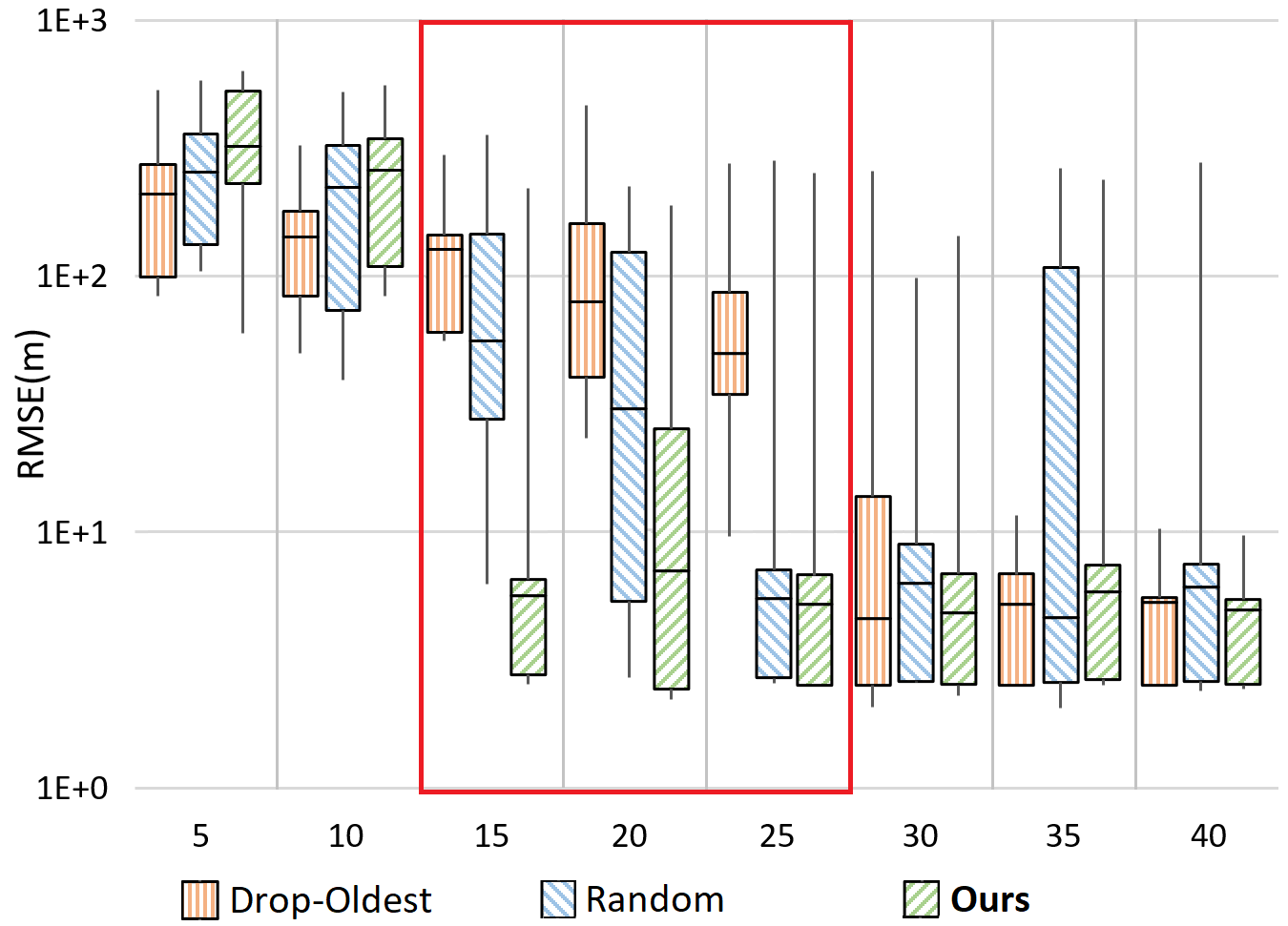}
\caption{The effect of buffer size. Our ORBBuf method can tolerate the same level of network unreliability with a smaller buffer size.}
\label{fig:buf}
\end{figure}

We further test the effect of varying the buffer size.
We repeat the experiment using sequence No.00 of the KITTI dataset and the network trace labelled Car02 with different buffer sizes.
We repeat each test case 10 times, and the results are summarized in the box-plot figure in Figure~\ref{fig:buf}.
When using the Drop-Oldest method, the resulting RMSE becomes low when the buffer size increases to 30 or more.
When using our ORBBuf method, the resulting RMSE becomes low since the buffer size is 15 or more.
When using the Random method, the resulting RMSE is not stable, even when the buffer size is 35.
This result indicates that our ORBBuf method can tolerate the same level of network unreliability with a smaller buffer size.

\begin{figure}[htb]
\centering
\includegraphics[width=\columnwidth]{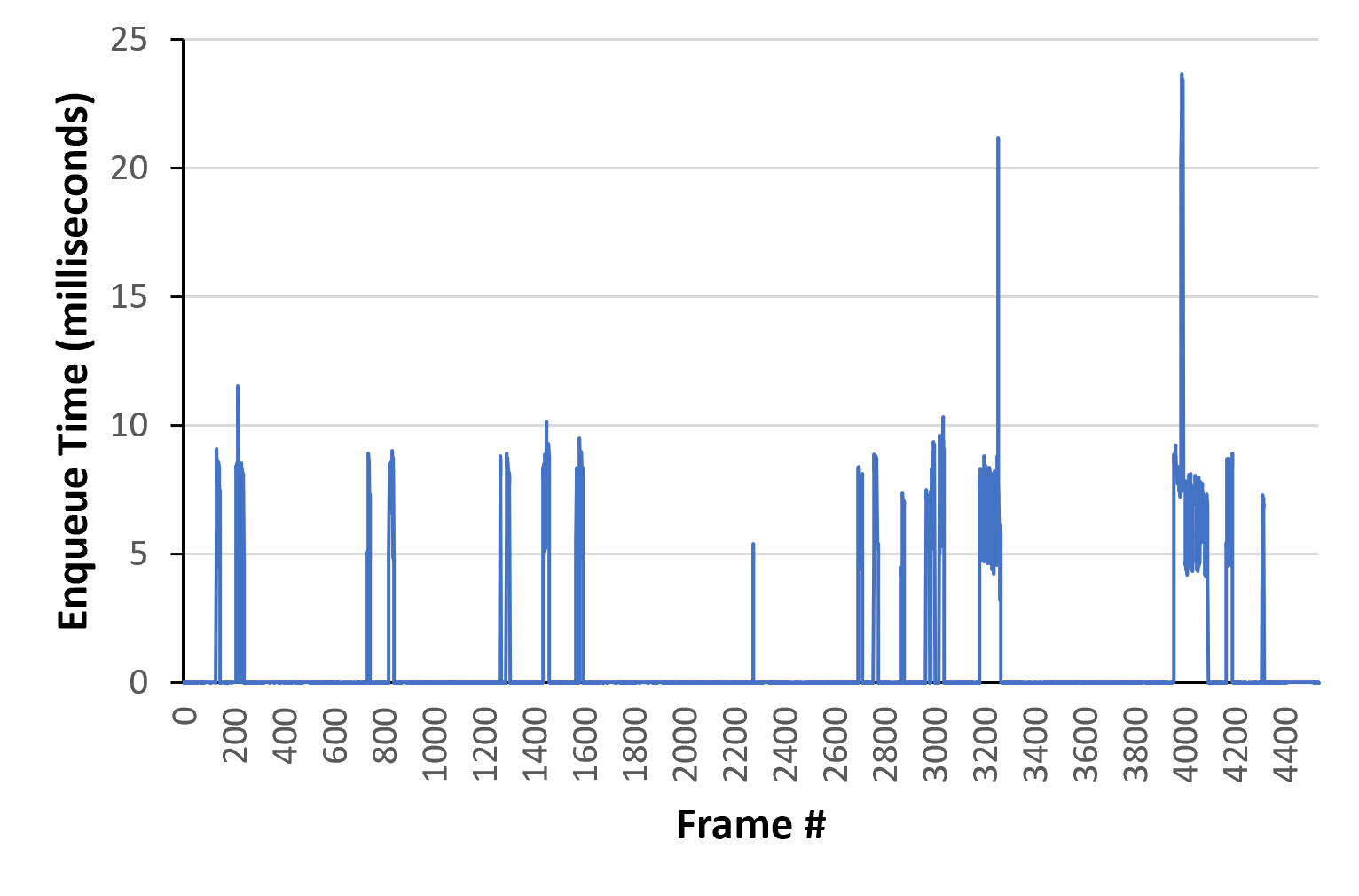}
\caption{Running time of our ORBBuf method. Our ORBBuf method introduces negligible time overhead.}
\label{fig:time}
\end{figure}

In addition, we test the running time of our ORBBuf method during the experiment.
Since our ORBBuf method introduces the calculation of the ORB feature into the message enqueuing routine, it introduces some time overhead.
We record the time to enqueue each frame during the previous experiment.
The result using sequence 00 of the KITTI dataset and the network trace labelled Car02 is shown in Figure~\ref{fig:time}.
This result is related to the used network trace.
When the network is stable, no time overhead is introduced at all.
When the network bandwidth is low or network interruption occurs, our ORBBuf method starts to calculate the ORB features.
In the worst case, our ORBBuf method introduces overhead of about 23ms, which does not affect the periodic data transmission (i.e. 10 Hz in this experiment and 25Hz in the next experiment).

\subsection{Real-World Network}

In this experiment, we build a Turtlebot3 Burger with a Raspberry Pi and 1GB memory.
Images captured by a camera on top of the Turtlebot are transmitted to the server via a public WiFi router.
We run the DSO algorithm on the server and use a Bluetooth keyboard to control the robot moving along the same path when using different buffering methods.

The results are shown in Figure~\ref{fig:feature}.
When using the Drop-Oldest and Random methods, the SLAM algorithm loses its trajectory.
When using our ORBBuf method, the SLAM algorithm successfully estimates the correct trajectory (the red curve) and builds a sparse 3D map (the white points).

%To make fair comparisons for different methods, the same buffer size is used for different methods and all data sequences are replayed multiple times.
%For each data sequence and each test round, we replay the sequence four times.
%The first time, the sequence is replayed at the server side, and %the result serves as the ground truth.
%The second time, the Drop-Oldest method is employed.
%The third time, the Random method is employed.
%The last time, our ORBBuf method is employed.

~

Overall, we have shown that our ORBBuf method can be used for different kinds of network situations and can adapt different kinds of input sensor data.
The network interruptions indeed affect the remote SLAM systems, and after using our ORBBuf method, SLAM systems become more robust against network unreliability.

\section{Conclusion}\label{sec:conclusion}

We present a novel buffering method for remote visual SLAM systems.
By considering the similarity between frames inside the buffer, our proposed buffering method helps overcome the network interruptions.
We integrated our ORBBuf method with ROS, a commonly used communication middleware.
Experimental results demonstrated that our ORBBuf method helps visual SLAM algorithms be more robust against network unreliability and reduces the RMSE up to 50 times comparing with the Drop-Oldest and Random buffering methods.

%Overall, in order to resist network unreliability, our work has taken the correlations between visual frames into account in the buffering method.
%It is also promising to further take the motion plan of the robot into account.

\bibliographystyle{IEEEtran}
\bibliography{main}

\begin{thebibliography}{10}
\providecommand{\url}[1]{#1}
\csname url@rmstyle\endcsname
\providecommand{\newblock}{\relax}
\providecommand{\bibinfo}[2]{#2}
\providecommand\BIBentrySTDinterwordspacing{\spaceskip=0pt\relax}
\providecommand\BIBentryALTinterwordstretchfactor{4}
\providecommand\BIBentryALTinterwordspacing{\spaceskip=\fontdimen2\font plus
\BIBentryALTinterwordstretchfactor\fontdimen3\font minus
  \fontdimen4\font\relax}
\providecommand\BIBforeignlanguage[2]{{%
\expandafter\ifx\csname l@#1\endcsname\relax
\typeout{** WARNING: IEEEtran.bst: No hyphenation pattern has been}%
\typeout{** loaded for the language `#1'. Using the pattern for}%
\typeout{** the default language instead.}%
\else
\language=\csname l@#1\endcsname
\fi
#2}}

\bibitem{DBLP:journals/ijrr/WhelanKJFLM15}
T.~Whelan, M.~Kaess, H.~Johannsson, M.~F. Fallon, J.~J. Leonard, and
  J.~McDonald, ``Real-time large-scale dense {RGB-D} {SLAM} with volumetric
  fusion,'' \emph{I. J. Robotics Res.}, vol.~34, no. 4-5, pp. 598--626, 2015.

\bibitem{DBLP:conf/cvpr/ChoiZK15}
S.~Choi, Q.~Zhou, and V.~Koltun, ``Robust reconstruction of indoor scenes,'' in
  \emph{{IEEE} Conference on Computer Vision and Pattern Recognition, {CVPR}
  2015, Boston, MA, USA, June 7-12, 2015}, 2015, pp. 5556--5565.

\bibitem{DBLP:conf/mm/ZhangWLH19}
H.~Zhang, G.~Wang, Z.~Lei, and J.~Hwang, ``Eye in the sky: Drone-based object
  tracking and 3d localization,'' in \emph{Proceedings of the 27th {ACM}
  International Conference on Multimedia, {MM} 2019, Nice, France, October
  21-25, 2019}, L.~Amsaleg, B.~Huet, M.~A. Larson, G.~Gravier, H.~Hung, C.~Ngo,
  and W.~T. Ooi, Eds.\hskip 1em plus 0.5em minus 0.4em\relax {ACM}, 2019, pp.
  899--907.

\bibitem{DBLP:conf/icra/OpdenboschOGAS18}
D.~V. Opdenbosch, M.~Oelsch, A.~Garcea, T.~Aykut, and E.~G. Steinbach,
  ``Selection and compression of local binary features for remote visual
  {SLAM},'' in \emph{2018 {IEEE} International Conference on Robotics and
  Automation, {ICRA} 2018, Brisbane, Australia, May 21-25, 2018}.\hskip 1em
  plus 0.5em minus 0.4em\relax {IEEE}, 2018, pp. 7270--7277.

\bibitem{DBLP:journals/tog/Dong0ZTXNC19}
S.~Dong, K.~Xu, Q.~Zhou, A.~Tagliasacchi, S.~Xin, M.~Nie{\ss}ner, and B.~Chen,
  ``Multi-robot collaborative dense scene reconstruction,'' \emph{{ACM} Trans.
  Graph.}, vol.~38, no.~4, pp. 84:1--84:16, 2019.

\bibitem{DBLP:conf/icra/JamiesonHG20}
S.~Jamieson, J.~P. How, and Y.~A. Girdhar, ``Active reward learning for
  co-robotic vision based exploration in bandwidth limited environments,'' in
  \emph{2020 {IEEE} International Conference on Robotics and Automation, {ICRA}
  2020, Paris, France, May 31 - August 31, 2020}.\hskip 1em plus 0.5em minus
  0.4em\relax {IEEE}, 2020, pp. 1806--1812.

\bibitem{DBLP:journals/vrih/ZouTY19}
D.~Zou, P.~Tan, and W.~Yu, ``Collaborative visual {SLAM} for multiple agents:
  {A} brief survey,'' \emph{Virtual Real. Intell. Hardw.}, vol.~1, no.~5, pp.
  461--482, 2019.

\bibitem{ros}
M.~Quigley, B.~Gerkey, K.~Conley, J.~Faust, T.~Foote, J.~Leibs, E.~Berger,
  R.~Wheeler, and A.~Ng, ``Ros: an open-source robot operating system,'' in
  \emph{{IEEE} International Conference on Robotics and Automation, {ICRA},
  Workshop on Open Source Software, Kobe, Japan}, 2009.

\bibitem{DBLP:conf/mass/EzifeLY17}
F.~Ezife, W.~Li, and S.~Yang, ``A survey of buffer management strategies in
  delay tolerant networks,'' in \emph{14th {IEEE} International Conference on
  Mobile Ad Hoc and Sensor Systems, {MASS} 2017, Orlando, FL, USA, October
  22-25, 2017}.\hskip 1em plus 0.5em minus 0.4em\relax {IEEE} Computer Society,
  2017, pp. 599--603.

\bibitem{DBLP:journals/jfr/GTSL16}
S.~Saeedi, M.~Trentini, M.~Seto, and H.~Li, ``Multiple-robot simultaneous
  localization and mapping: {A} review,'' \emph{J. Field Robotics}, vol.~33,
  no.~1, pp. 3--46, 2016.

\bibitem{DBLP:conf/egh/HasselgrenA06}
J.~Hasselgren and T.~Akenine{-}M{\"{o}}ller, ``Efficient depth buffer
  compression,'' in \emph{Proceedings of the 21st {ACM} {SIGGRAPH/EUROGRAPHICS}
  symposium on Graphics hardware, Vienna, Austria, September 3-4, 2006}, 2006,
  pp. 103--110.

\bibitem{DBLP:journals/tip/ChenHC18}
J.~Chen, J.~Hou, and L.~Chau, ``Light field compression with disparity-guided
  sparse coding based on structural key views,'' \emph{{IEEE} Trans. Image
  Processing}, vol.~27, no.~1, pp. 314--324, 2018.

\bibitem{DBLP:conf/icip/KimH07}
S.~Kim and Y.~Ho, ``Mesh-based depth coding for 3d video using hierarchical
  decomposition of depth maps,'' in \emph{Proceedings of the International
  Conference on Image Processing, {ICIP} 2007, September 16-19, 2007, San
  Antonio, Texas, {USA}}, 2007, pp. 117--120.

\bibitem{DBLP:journals/ral/KahlerPVM16}
O.~K{\"{a}}hler, V.~A. Prisacariu, J.~P.~C. Valentin, and D.~W. Murray,
  ``Hierarchical voxel block hashing for efficient integration of depth
  images,'' \emph{{IEEE} Robotics and Automation Letters}, vol.~1, no.~1, pp.
  192--197, 2016.

\bibitem{DBLP:journals/tvcg/GolodetzCLPMT18}
S.~Golodetz, T.~Cavallari, N.~A. Lord, V.~A. Prisacariu, D.~W. Murray, and
  P.~H.~S. Torr, ``Collaborative large-scale dense 3d reconstruction with
  online inter-agent pose optimisation,'' \emph{{IEEE} Trans. Vis. Comput.
  Graph.}, vol.~24, no.~11, pp. 2895--2905, 2018.

\bibitem{DBLP:journals/icl/HooftPWHRBT16}
J.~van~der Hooft, S.~Petrangeli, T.~Wauters, R.~Huysegems, P.~Rondao{-}Alface,
  T.~Bostoen, and F.~D. Turck, ``Http/2-based adaptive streaming of {HEVC}
  video over 4g/lte networks,'' \emph{{IEEE} Communications Letters}, vol.~20,
  no.~11, pp. 2177--2180, 2016.

\bibitem{DBLP:journals/twc/ChenGN19}
J.~Chen, X.~Ge, and Q.~Ni, ``Coverage and handoff analysis of 5g fractal small
  cell networks,'' \emph{{IEEE} Trans. Wireless Communications}, vol.~18,
  no.~2, pp. 1263--1276, 2019.

\bibitem{DBLP:journals/pami/EngelKC18}
J.~Engel, V.~Koltun, and D.~Cremers, ``Direct sparse odometry,'' \emph{{IEEE}
  Trans. Pattern Anal. Mach. Intell.}, vol.~40, no.~3, pp. 611--625, 2018.

\bibitem{DBLP:journals/trob/QinLS18}
T.~Qin, P.~Li, and S.~Shen, ``Vins-mono: {A} robust and versatile monocular
  visual-inertial state estimator,'' \emph{{IEEE} Trans. Robotics}, vol.~34,
  no.~4, pp. 1004--1020, 2018.

\bibitem{DBLP:journals/ijrr/GeigerLSU13}
A.~Geiger, P.~Lenz, C.~Stiller, and R.~Urtasun, ``Vision meets robotics: The
  {KITTI} dataset,'' \emph{I. J. Robotics Res.}, vol.~32, no.~11, pp.
  1231--1237, 2013.

\bibitem{DBLP:journals/corr/EngelUC16}
\BIBentryALTinterwordspacing
J.~Engel, V.~C. Usenko, and D.~Cremers, ``A photometrically calibrated
  benchmark for monocular visual odometry,'' \emph{CoRR}, vol. abs/1607.02555,
  2016. [Online]. Available: \url{http://arxiv.org/abs/1607.02555}
\BIBentrySTDinterwordspacing

\bibitem{DBLP:journals/trob/Mur-ArtalMT15}
R.~Mur{-}Artal, J.~M.~M. Montiel, and J.~D. Tard{\'{o}}s, ``{ORB-SLAM:} {A}
  versatile and accurate monocular {SLAM} system,'' \emph{{IEEE} Trans.
  Robotics}, vol.~31, no.~5, pp. 1147--1163, 2015.

\bibitem{DBLP:journals/trob/Mur-ArtalT17}
R.~Mur{-}Artal and J.~D. Tard{\'{o}}s, ``{ORB-SLAM2:} an open-source {SLAM}
  system for monocular, stereo, and {RGB-D} cameras,'' \emph{{IEEE} Trans.
  Robotics}, vol.~33, no.~5, pp. 1255--1262, 2017.

\bibitem{DBLP:journals/corr/abs-2007-11898}
\BIBentryALTinterwordspacing
C.~Campos, R.~Elvira, J.~J.~G. Rodr{\'{\i}}guez, J.~M.~M. Montiel, and J.~D.
  Tard{\'{o}}s, ``{ORB-SLAM3:} an accurate open-source library for visual,
  visual-inertial and multi-map {SLAM},'' \emph{CoRR}, vol. abs/2007.11898,
  2020. [Online]. Available: \url{https://arxiv.org/abs/2007.11898}
\BIBentrySTDinterwordspacing

\bibitem{DBLP:conf/iccv/RubleeRKB11}
E.~Rublee, V.~Rabaud, K.~Konolige, and G.~R. Bradski, ``{ORB:} an efficient
  alternative to {SIFT} or {SURF},'' in \emph{{IEEE} International Conference
  on Computer Vision, {ICCV} 2011, Barcelona, Spain, November 6-13, 2011},
  D.~N. Metaxas, L.~Quan, A.~Sanfeliu, and L.~V. Gool, Eds.\hskip 1em plus
  0.5em minus 0.4em\relax {IEEE} Computer Society, 2011, pp. 2564--2571.

\bibitem{DBLP:journals/trob/CadenaCCLSN0L16}
C.~Cadena, L.~Carlone, H.~Carrillo, Y.~Latif, D.~Scaramuzza, J.~Neira, I.~D.
  Reid, and J.~J. Leonard, ``Past, present, and future of simultaneous
  localization and mapping: Toward the robust-perception age,'' \emph{{IEEE}
  Trans. Robotics}, vol.~32, no.~6, pp. 1309--1332, 2016.

\bibitem{DBLP:journals/jfr/LabbeM19}
M.~Labb{\'{e}} and F.~Michaud, ``Rtab-map as an open-source lidar and visual
  simultaneous localization and mapping library for large-scale and long-term
  online operation,'' \emph{J. Field Robotics}, vol.~36, no.~2, pp. 416--446,
  2019.

\bibitem{DBLP:conf/iccv/NewcombeLD11}
R.~A. Newcombe, S.~Lovegrove, and A.~J. Davison, ``{DTAM:} dense tracking and
  mapping in real-time,'' in \emph{{IEEE} International Conference on Computer
  Vision, {ICCV} 2011, Barcelona, Spain, November 6-13, 2011}, D.~N. Metaxas,
  L.~Quan, A.~Sanfeliu, and L.~V. Gool, Eds.\hskip 1em plus 0.5em minus
  0.4em\relax {IEEE} Computer Society, 2011, pp. 2320--2327.

\bibitem{DBLP:conf/eccv/EngelSC14}
J.~Engel, T.~Sch{\"{o}}ps, and D.~Cremers, ``{LSD-SLAM:} large-scale direct
  monocular {SLAM},'' in \emph{Computer Vision - {ECCV} 2014 - 13th European
  Conference, Zurich, Switzerland, September 6-12, 2014, Proceedings, Part
  {II}}, ser. Lecture Notes in Computer Science, D.~J. Fleet, T.~Pajdla,
  B.~Schiele, and T.~Tuytelaars, Eds., vol. 8690.\hskip 1em plus 0.5em minus
  0.4em\relax Springer, 2014, pp. 834--849.

\bibitem{DBLP:conf/iswc/CastleKM08}
R.~O. Castle, G.~Klein, and D.~W. Murray, ``Video-rate localization in multiple
  maps for wearable augmented reality,'' in \emph{12th {IEEE} International
  Symposium on Wearable Computers {(ISWC} 2008), September 28 - October 1,
  2008, Pittsburgh, PA, {USA}}.\hskip 1em plus 0.5em minus 0.4em\relax {IEEE}
  Computer Society, 2008, pp. 15--22.

\bibitem{DBLP:journals/ras/RiazueloCM14}
L.~Riazuelo, J.~Civera, and J.~M.~M. Montiel, ``C\({}^{\mbox{2}}\)tam: {A}
  cloud framework for cooperative tracking and mapping,'' \emph{Robotics Auton.
  Syst.}, vol.~62, no.~4, pp. 401--413, 2014.

\bibitem{DBLP:journals/ral/KarrerSC18}
M.~Karrer, P.~Schmuck, and M.~Chli, ``{CVI-SLAM} - collaborative
  visual-inertial {SLAM},'' \emph{{IEEE} Robotics Autom. Lett.}, vol.~3, no.~4,
  pp. 2762--2769, 2018.

\bibitem{DBLP:conf/icra/CieslewskiCS18}
T.~Cieslewski, S.~Choudhary, and D.~Scaramuzza, ``Data-efficient decentralized
  visual {SLAM},'' in \emph{2018 {IEEE} International Conference on Robotics
  and Automation, {ICRA} 2018, Brisbane, Australia, May 21-25, 2018}.\hskip 1em
  plus 0.5em minus 0.4em\relax {IEEE}, 2018, pp. 2466--2473.

\bibitem{DBLP:conf/icra/KimKFLBRT10}
B.~Kim, M.~Kaess, L.~Fletcher, J.~J. Leonard, A.~Bachrach, N.~Roy, and S.~J.
  Teller, ``Multiple relative pose graphs for robust cooperative mapping,'' in
  \emph{{IEEE} International Conference on Robotics and Automation, {ICRA}
  2010, Anchorage, Alaska, USA, 3-7 May 2010}.\hskip 1em plus 0.5em minus
  0.4em\relax {IEEE}, 2010, pp. 3185--3192.

\bibitem{DBLP:journals/ral/LajoieRCCB20}
P.~Lajoie, B.~Ramtoula, Y.~Chang, L.~Carlone, and G.~Beltrame, ``{DOOR-SLAM:}
  distributed, online, and outlier resilient {SLAM} for robotic teams,''
  \emph{{IEEE} Robotics Autom. Lett.}, vol.~5, no.~2, pp. 1656--1663, 2020.

\bibitem{DBLP:conf/icra/CarloneNDBI10}
L.~Carlone, M.~E.~K. Ng, J.~Du, B.~Bona, and M.~Indri, ``Rao-blackwellized
  particle filters multi robot {SLAM} with unknown initial correspondences and
  limited communication,'' in \emph{{IEEE} International Conference on Robotics
  and Automation, {ICRA} 2010, Anchorage, Alaska, USA, 3-7 May 2010}, 2010, pp.
  243--249.

\bibitem{DBLP:conf/mmsys/Stockhammer11}
T.~Stockhammer, ``Dynamic adaptive streaming over {HTTP} -: standards and
  design principles,'' in \emph{Proceedings of the Second Annual {ACM} {SIGMM}
  Conference on Multimedia Systems, MMSys 2011, Santa Clara, CA, USA, February
  23-25, 2011}, A.~C. Begen and K.~Mayer{-}Patel, Eds.\hskip 1em plus 0.5em
  minus 0.4em\relax {ACM}, 2011, pp. 133--144.

\bibitem{DBLP:journals/sigact/Goldwasser10}
M.~H. Goldwasser, ``A survey of buffer management policies for packet
  switches,'' \emph{{SIGACT} News}, vol.~41, no.~1, pp. 100--128, 2010.

\bibitem{DBLP:conf/secon/KrifaBS08}
A.~Krifa, C.~Barakat, and T.~Spyropoulos, ``Optimal buffer management policies
  for delay tolerant networks,'' in \emph{Proceedings of the Fifth Annual
  {IEEE} Communications Society Conference on Sensor, Mesh and Ad Hoc
  Communications and Networks, {SECON} 2008, June 16-20, 2008, Crowne Plaza,
  San Francisco International Airport, California, {USA}}, 2008, pp. 260--268.

\bibitem{DBLP:conf/csa2/KimW17}
S.~Kim and Y.~Won, ``Frame rate control buffer management technique for
  high-quality real-time video conferencing system,'' in \emph{Advances in
  Computer Science and Ubiquitous Computing - {CSA/CUTE} 2017, Taichung,
  Taiwan, 18-20 December.}, 2017, pp. 777--783.

\bibitem{DBLP:journals/cn/Scalosub17}
G.~Scalosub, ``Towards optimal buffer management for streams with packet
  dependencies,'' \emph{Computer Networks}, vol. 129, pp. 207--214, 2017.

\bibitem{DBLP:journals/pomacs/YangWH17}
L.~Yang, W.~S. Wong, and M.~H. Hajiesmaili, ``An optimal randomized online
  algorithm for qos buffer management,'' \emph{{POMACS}}, vol.~1, no.~2, pp.
  36:1--36:26, 2017.

\bibitem{DBLP:conf/apcc/KimuraM17}
T.~Kimura and M.~Muraguchi, ``Buffer management policy based on message rarity
  for store-carry-forward routing,'' in \emph{23rd Asia-Pacific Conference on
  Communications, {APCC} 2017, Perth, Australia, December 11-13, 2017}, 2017,
  pp. 1--6.

\bibitem{quic}
T.~C. Projects, ``Quic, a multiplexed stream transport over udp,''
  \url{https://www.chromium.org/quic}, 2019.

\bibitem{ros_comm}
ROS, ``ros\_comm project,'' \url{https://github.com/ros/ros_comm}, 2019.

\bibitem{DBLP:conf/iros/QinS18}
T.~Qin and S.~Shen, ``Online temporal calibration for monocular visual-inertial
  systems,'' in \emph{2018 {IEEE/RSJ} International Conference on Intelligent
  Robots and Systems, {IROS} 2018, Madrid, Spain, October 1-5, 2018}, 2018, pp.
  3662--3669.

\bibitem{grupp2017evo}
M.~Grupp, ``evo: Python package for the evaluation of odometry and slam.''
  \url{https://github.com/MichaelGrupp/evo}, 2017.

\end{thebibliography}

\end{document}